\documentclass[runningheads]{llncs}
\usepackage[T1]{fontenc}
\usepackage{graphicx}
\usepackage{booktabs}
\usepackage{amsmath}
\usepackage[misc]{ifsym}

\usepackage{mwe}
\usepackage{hyperref}
\usepackage{orcidlink}
\usepackage[colorinlistoftodos]{todonotes}

\begin{document}

\title{Everyone deserves their voice to be heard: Analyzing Predictive Gender Bias in ASR Models Applied to Dutch Speech Data} 

\titlerunning{Analyzing Predictive Gender Bias in Dutch ASR}

\author{Rik Raes\inst{1} \orcidlink{https://orcid.org/0009-0002-9647-6072} \and
Saskia E. Lensink\inst{2} \orcidlink{https://orcid.org/0000-0002-5395-1008} \and
Mykola Pechenizkiy\inst{1}\orcidlink{https://orcid.org/0000-0003-4955-0743}}

\authorrunning{R. Raes et al.}

\institute{Eindhoven University of Technology, Eindhoven, The Netherlands \and
Netherlands Organisation for Applied Scientific Research (TNO), The Hague, The Netherlands \\ 
}

\maketitle              

\begin{abstract}
Recent research has shown that state-of-the-art (SotA) Automatic Speech Recognition (ASR) systems, such as Whisper, often exhibit predictive biases that disproportionately affect various demographic groups. This study focuses on identifying the performance disparities of Whisper models on Dutch speech data from the Common Voice dataset and the Dutch National Public Broadcasting organisation. We analyzed the word error rate, character error rate and a BERT-based semantic similarity across gender groups. We used the moral framework of Weerts et al. (2022) to assess quality of service harms and fairness, and to provide a nuanced discussion on the implications of these biases, particularly for automatic subtitling. Our findings reveal substantial disparities in word error rate (WER) among gender groups across all model sizes, with bias identified through statistical testing.

\end{abstract}

\section{Introduction}
Automatic Speech Recognition (ASR) technology has increasingly become integral to a wide range of applications, enhancing user interaction through virtual assistants and making multimedia content more accessible via automatic subtitling. 
Recent studies have substantially advanced the accuracy of ASR systems in transcribing human speech. However, research consistently reveals that these systems often manifest performance disparities that correlate with demographic variables. 
Biases in ASR systems can have profound implications, potentially exacerbating social inequalities and raising serious ethical concerns in high-stakes applications like employment, where ASR technology is used in tools for automated video interviews \cite{hovy2016social,shahani2015now,morrison2017speech,ajunwa2016hiring}. This research aims to address these substantial issues by employing a rigorous, morally justified framework to evaluate biases and fairness within Whisper, a state-of-the-art ASR model \cite{whisperai}, specifically analyzing its performance with Dutch language data.

Previous research has shown performance differences in the recognition rate of ASR models due to speaker variations in demographic attributes. Some works show better recognition for male speech \cite{tatman2017gender,garnerin2019gender,garnerin2021investigating,meyer2020artie}, while most indicate better performance for female speech \cite{feng2021quantifying,sawalha2013effects,moro2019study,adda2005speech,liu2022towards,feng2023towards,fuckner2023uncovering} and other report mixed findings or no (significant) differences \cite{tatman2017effects,liu2022model,chan2022training}. Age-related disparities also surface, with studies generally showing superior ASR performance for teenagers over children \cite{feng2021quantifying,feng2023towards,fuckner2023uncovering}. Accent and native language further influence ASR performance, with native speakers or those who acquired the language early typically experiencing better recognition rates compared to non-native speakers or those with heavy accents \cite{chan2022training,zhangcomparing,zhang2022mitigating,feng2023towards,fuckner2023uncovering,palanica2019you,tatman2017effects}.

Despite the widespread acknowledgment of biases, the methodological rigor in evaluating such biases often lacks. A substantial portion of existing studies use statistical tests such as ANOVA, Kruskal-Wallis, and linear mixed models to analyze performance disparities \cite{suresh2021framework,chan2022training,garnerin2019gender,fuckner2023uncovering,meyer2020artie,feng2023towards}. However, these studies rarely discuss the validity of the underlying assumptions of these tests, which could potentially lead to misleading conclusions. This is one of the indications that a more rigorous methodology still needs to be established in the ASR (predictive) bias research. 
Existing work also typically does not provide moral justification regarding the usage of metrics to measure and argue about fairness. 
As noted by Blodgett et al., the techniques used to measure fairness and biases must be justified and reasoned, matching with the motivations \cite{blodgett2020language}. Furthermore, the authors propose that future research should ``Provide explicit statements of why the system behaviors that are described as `bias' are harmful, in what ways, and to whom. Be forthright about the normative reasoning (Green, 2019) underlying these statements." \cite{blodgett2020language}. Therefore, simply coming up with a metric a model must abide by is not always morally justified. Highlighting the mathematical properties does not vindicate using a fairness metric without considering the moral implications, assumptions, or considerations \cite{hertweck2021moral}. 
This study makes first steps to address the lack of moral justification by integrating moral considerations into the evaluation of ASR systems, moving beyond the mere identification of biases to encompass a broader discussion on ethical implications and fairness \cite{weerts2022does}. 


Most existing research on ASR bias focuses on English data, leaving performance across other languages and demographic settings less explored. We use data from both Mozilla's Common Voice dataset and a dataset comprising Dutch TV and radio shows from the NPO, the Dutch National Public Broadcasting organisation (\textit{Nederlandse Publieke Omroep}). Our study aims to provide a comprehensive analysis of the Whisper ASR model's performance across a range of speech types and gender groups, highlighting both its capabilities and limitations. This thorough evaluation helps identify potential areas for improvement in ASR technology, ensuring it delivers equitable and effective service across all user groups.

The rest of the paper details our methodology, presents findings on gender bias in Whisper models, interprets these findings within our moral framework, and concludes with insights and recommendations for future research.



\section{Methods}
\subsection{Whisper ASR}
Whisper, a recent ASR breakthrough, employs a zero-shot foundational model trained on a vast dataset (680k hours) sourced exclusively from human-generated content on the internet \cite{whisperai,ghorbani2021scaling}. This model utilizes an encoder-decoder Transformer architecture, processing inputs through an 80-channel log-magnitude Mel spectrogram to optimize transcription, translation, and language identification across multiple languages. Notably, the model's architecture facilitates handling diverse audio qualities by resampling all inputs to 16,000 Hz and normalizing them. 
OpenAI has made Whisper’s code and models public, encouraging further research and optimization. 

There exist multiple variants of the Whisper model, differing in size and model complexity. This research deploys and evaluates the performance of five different Whisper models: tiny (39M parameters), base (74M parameters), small (244M parameters), medium (769M parameters), and large (1550M parameters). The largest Whisper model is said to approach human-level accuracy and robustness in English speech recognition, and smaller models also perform outstandingly, making them attractive for many organizations to use, analyze, and improve for their applications. However, these foundational Whisper models have only been introduced recently. Therefore, there has yet to be an extensive analysis of the possible demographic biases of these models, what harms they pose, and how we can effectively measure them, not only from a statistical perspective but also from a moral point of view.


\subsection{Datasets} 
Our study utilizes two primary sources of data to assess the performance of Whisper, an advanced ASR model developed by OpenAI. The first dataset is the Common Voice dataset, an open-source multi-language collection curated by Mozilla \cite{ardila2019common}. It consists of audio clips where speakers read single sentences, enhanced with metadata containing demographic information such as gender, age, and accent. Each instance is verified by at least two users to ensure accuracy. Specifically, we used the $13^{th}$ edition of the Dutch Common Voice dataset, which offers a rich source of varied speech data. This dataset is crucial for assessing quality-of-service harms in ASR systems, particularly by evaluating transcription accuracy against the similarity in meaning between predicted and ground truth transcriptions. Despite some skewed distributions, such as a predominance of the "Nederlands Nederlands" accent and a higher number of male instances, there are ample data points for robust analysis. However, groups with fewer instances are considered unrepresentative and will be excluded from this research, ensuring a comprehensive and representative evaluation of biases within ASR systems. The dataset contains a total of 86,798 fragments, with 71,097 instances having demographic attributes such as age (50,853), gender (50,518), or accent (66,185).

The second dataset consists of almost 37 hours of transcribed data from the Dutch National Public Broadcasting organisation, containing 62 TV shows and four radio shows with word-by-word and speaker-segmented transcriptions. This dataset has been composed to provide the broadest possible representation of the types of programs produced by the NPO. It includes various programs such as political debates, comedy shows, and children's programs, offering a mix of read and spontaneous speech. We categorize these shows into three types: eloquent (political shows and news programs), non-eloquent (comedy, children's programs, informal talk shows), and radio shows, each presenting different transcription challenges. Segments in this data vary in length from single words to multiple paragraphs. Gender metadata is inferred from available contextual clues, such as speaker names. 
We expect to see better performances on TV shows containing eloquent speech than on more informal speech from both TV and radio. 

\subsection{Measuring Fairness: a Moral Framework}
To discuss all relevant moral implications for deducing our fairness evaluation framework, we first need to identify possible harms in the context of automatic subtitling. Following Crawford \cite{nips2017crawford}, we have identified quality-of-service harm to be most relevant here. Quality-of-service harm refers to a skewed distribution of errors across groups, impacting the accuracy of subtitles, which are essential for understanding across language barriers, and can influence public perception of the speakers. Inaccuracies in ASR might advantage or disadvantage individuals or groups, influencing public engagement and competitive standing.

In media, inaccurate subtitles can distort public perception and understanding, leading to misrepresentations of individuals and groups featured in content. This can have far-reaching consequences, affecting how these groups are viewed by society at large. In educational settings, biases in subtitle accuracy could hinder comprehension and learning, particularly for non-native speakers and the deaf or hard-of-hearing community. This not only affects individual learning outcomes but also perpetuates systemic inequalities by providing unequal access to educational resources. The exclusive nature of biased ASR technology undermines efforts to create inclusive environments where all students have equal opportunities to succeed. Furthermore, worse performance for groups that already feel marginalized, such as accented speakers, exacerbates their sense of exclusion and increases quality-of-service harms for those already experiencing inequalities. Addressing these biases is essential to ensure that ASR technology contributes to greater social equity rather than reinforcing existing disparities.

Our research employs a moral reasoning framework to assess the fairness of ASR systems, focusing on quality-of-service harms linked to transcription accuracy. The framework, extended by Weerts et al. \cite{weerts2022does} from Hertweck et al.'s work \cite{hertweck2021moral}, consists of five spaces: Potential, Construct, Observation, Decision, and Utility. These spaces help identify different biases, such as life's bias, measurement bias, and technical bias, by examining the distortions between each space. This structured approach ensures a comprehensive understanding of how biases occur and how they affect different demographic groups.

The Potential Space represents innate abilities influencing vocal characteristics, while the Construct Space reflects actual vocal traits. The Observation Space comprises the audio data, the Decision Space includes ASR transcriptions, and the Utility Space denotes the benefits or harms experienced by individuals based on these transcriptions. By focusing on the Construct Space for utility considerations, the framework aims to capture both innate and chosen aspects of speech, promoting a fair and inclusive ASR system.

In applying this framework, we aim to identify and quantify technical bias, particularly the distortions between the Observed and Decision Spaces. This involves analyzing predictive biases and ensuring that the ASR systems provide equitable utility across different demographic groups. In order to perform this utility evaluation we proprose to use a WER Parity metric, which will be detailed in the next section. We focus on both statistical and semantic aspects of predictive performance, since the quality or accuracy of transcriptions in ASR depends on how close sentences are to each other but also on their similarity in meaning.

\subsection{Analytical Techniques and Benchmarking Methods}

Our evaluation framework includes computing the Word Error Rate (WER) and Character Error Rate (CER), fundamental metrics in ASR research. WER (Eq. \ref{eq:wer}) measures word-level accuracy, while CER (Eq. \ref{eq:cer}) provides character-level insights, addressing WER's limitations such as its unboundedness and asymmetry for insertions and deletions \cite{errattahi2018automatic}. 

\begin{equation} \label{eq:cer}
    CER = \frac{S_{c} + D_{c} + I_{c}}{N_{c}}
\end{equation}
\begin{equation} \label{eq:wer}
    WER = \frac{S_{w} + D_{w} + I_{w}}{N_{w}}
\end{equation}

\begin{center}
    , where S = \#substitutions \\ \hspace{.5cm} D = \#deletions \\ \hspace{.5cm} I = \#insertions \\ \hspace{2.5cm} N = \#words in the reference
\end{center}

We not only measured a word- and character-based similarity metric, but also looked at a way to measure the similarity in terms of the overall meaning of the ground-truth transcripts and the model's output. This semantic similarity can be measured using a vector-based distance metric, which reflects a meaning-based measure of overlap between the ground-truth transcript and the ASR model's predictions. Specifically, we compute the BERT-based similarity (BSS) by obtaining embeddings of the ground-truth and predicted transcripts, on both the Common Voice and NPO data sets, using the BERT model (Bidirectional Encoder Representations from Transformers \cite{devlin2018bert}) and compute the cosine-distance between these embeddings (\ref{eq:bert_based_similarity}). Semantic similarity metrics have been proven to be simple and effective ways to measure how close two words or sentences are in meaning \cite{korenius2007principal,chakrabarti2000data}, but have to our knowledge not been used extensively in evaluating the quality of ASR transcripts. 

\begin{equation} \label{eq:bert_based_similarity}
    \begin{aligned}
    BSS(\hat{y_{i}}, y_{i}) = \\ cosine\_similarity(BERT(\hat{y_{i}}), BERT(y_{i})) = \\ (BERT(\hat{y_{i}}) \cdot BERT(y_{i})) / (||BERT(\hat{y_{i}})|| * ||BERT(y_{i})||)
    \end{aligned}
\end{equation}

Additionally, we consider model speed, crucial for real-time subtitling applications, and propose a fairness metric based on WER to analyze quality-of-service differences where statistical testing will identify predictive biases.

For fairness evaluation, we need to assess the amount of performance equality across groups. For this, we cannot solely rely on BERT-based similarity metrics. While the BSS metric provides unique insights into potential harms by considering semantics, it has been shown to contain biases in its embeddings \cite{kurita2019quantifying,bartl2020unmasking,bhardwaj2021investigating}. Consequently, using BSS for fairness measurement would be contradictory, which is why we employ it solely to analyze similarity. On the other hand, WER is the most widely used metric in ASR research and proves most valuable when combined with the proposed text standardization. Therefore, we propose a fairness metrics based on difference in word error rates: the WER Parity metric (Eq \ref{eq:mean_disparity_comb}).

\begin{equation} \label{eq:mean_disparity_comb}
    {\displaystyle \frac{\max(WER_{male},WER_{female})}{\min(WER_{male},WER_{female})} \leq (1 + \varepsilon)}
\end{equation}

The WER Parity metric evaluates whether WER differences between groups are within acceptable bounds, similar to the two one-sided t-test (TOST) procedures \cite{lakens2017equivalence}. While the exact optimal threshold was not determined a priori, 25\% was deemed reasonable after discussions with the NPO, given the need for a clear cutoff that could indicate unacceptable ASR performance differences between the gender. Future work could further refine this threshold value. If its outcome is True, then we pose that there are unfair and relevant differences between the groups, and if the outcome if False, then the outcomes are considered fair. This metric is focused on gender groups since this is the only attribute available in both datasets. However, it can be applied to other metrics if the fairness bounds are adjusted accordingly.

The fairness metric indicates whether differences in WER are relevant and, thus, unfair, whereas statistical tests will indicate whether a significant difference exists. In other words, if there is a significant difference following the statistical tests, it can be concluded that predictive bias is present in the underlying model on the relevant groups. The fairness metric yields additional insights about whether differences (predictive bias or not) are relevant and will have an disproportionate impact on different groups. This ensures a balanced approach to evaluating quality of service and potential biases and harms. Our comprehensive framework aims to identify and address predictive biases and quality-of-service disparities across different demographic groups, ensuring a fair and accurate evaluation of ASR systems.  

\begin{figure}[h!]
    \centering
    \includegraphics[width=\linewidth]{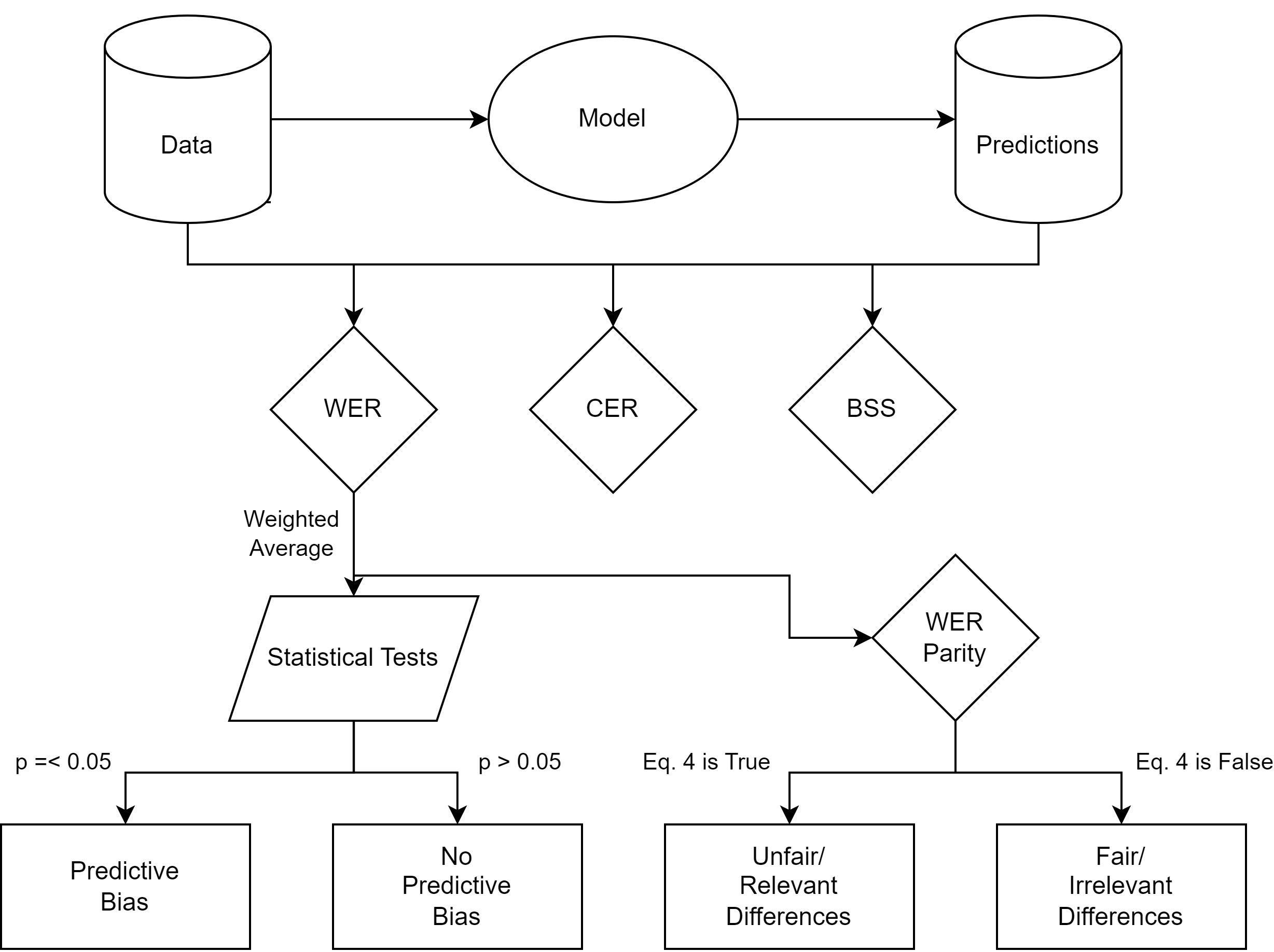}
    \caption{Flowchart of the evaluation methodology. Predictions are made by a model on a data set. The performance of this model is then evaluated by measuring the WER, CER, and BSS on all predictions with a weight according to the number of words in the instance. Next, the weighted average of these scores is taken by aggregating over the speaker IDs to remove data dependencies and perform statistical testing to find out whether predictive bias exists. Last, the models are evaluated on fairness using the WER Parity equation (\ref{eq:mean_disparity_comb}) on the weighted average WER scores.}
    \label{fig:eval_method}
\end{figure}

To ensure validity of evaluation, we apply text standardization to minimize non-semantic differences, using the basic text normalizer described in the original Whisper paper \cite{whisperai}. Furthermore, we use a weighted mean to aggregate performance metrics where the weight is equal to the number of words in an instance. This approach enhances robustness by reducing the impact of hallucinations and emphasizing context-rich instances, providing a more accurate assessment of Whisper's performance. Our framework ensures a more thorough evaluation of ASR systems. 
A high-level conceptual overview of the evaluation methodology is depicted in Figure \ref{fig:eval_method}.

We employed two-sample t-tests and one-way ANOVA for testing mean differences between groups. The analysis involves calculating per-speaker WER scores, using aggregation to remove inter-group dependence and enhance statistical power \cite{christie1990aggregation}. Assumptions for ANOVA and t-tests include normality, homogeneity of variance, and independence, tested with Shapiro-Wilk \cite{shapiro1965analysis} and Levene's tests. If assumptions are violated, alternatives like the Mann-Whitney U test and Welch ANOVA are employed. The study evaluates predictive biases and model performance differences using these statistical methods and a fairness metric. Group sizes indicate the use of t-tests for gender. 
The results highlight statistically significant predictive biases and inform further investigations into potential harms and quality of service disparities. 



\subsection{Experimental setup}
First, inference is performed with five Whisper models Tiny, Base, Small, Medium, and Large, using the Hugging Face Transformers library\footnote{https://huggingface.co/docs/transformers/index}. Audio clips are pre-processed like the original Whisper training process by re-sampling the audio to 16kHz and computing its log-Mel spectrogram. These input features are then used to generate output tokens with sequence-to-sequence mapping, after which the sequence of tokens are decoded into text tokens, which form the actual transcription.

Whisper models can only process 30-second audio chunks which works perfect for the sub 30-second instances in the Common Voice data. However, the dataset from NPO consists of lengthy audio programs. This requires either splitting the large audio files into sub-30-second segments or relying on Whisper's built-in long-form transcription strategy that uses a shifting 30-second audio context window. For effective predictions and better performance comparison, especially across gender groups, the data is segmented based on speakers, splitting audio into segments corresponding to each speaker's transcription timestamps. This segmentation method is crucial for accurate performance analysis across gender groups. If speaker segments exceed 30 seconds, a custom method can split these into the largest equal parts under 30 seconds, optimizing the model's context for each prediction. For example, a 42-second segment would be divided into two 21-second segments.

After performing inference, the WER, CER, and BSS metrics are computed on all instances, after which an aggregated comparison of mean scores per subgroup is made across the different gender groups. 
Next, significance tests are executed over all per-speaker WER scores to determine whether these differences are significant. The proposed \textit{WER Parity} fairness metric is then computed to reason about quality of service and significance tests over per-speaker WER scores to argue about predictive bias.

\section{Results}
\subsection{Common Voice Results}
In evaluating Whisper models on the Common Voice data set, we observe that larger models generally offer better performance across all metrics, albeit with slower inference times. Notably, these models tend to recognize female speech better than male speech as shown in Table \ref{tab:whisper_gender_pre_trained}. 

\begin{table}[h!]
\centering
\begin{tabular}{ll|l|l|l|l}
    \textbf{Model}          & \textbf{Subgroup}               & \textbf{Count (\#p)}          & \textbf{WER} & \textbf{CER} & \textbf{BSS}\\ \hline
    \textit{Tiny}           & \textit{Female}                 & 9588 (133)                    & 0.514	& 0.224 & 0.974                \\
                            & \textit{Male}                   & 40828 (502)                   & 0.498	& 0.207 & 0.974                \\ \hline
    \textit{Base}           & \textit{Female}                 & 9588 (133)                    & 0.340	& 0.135 & 0.974                \\
                            & \textit{Male}                   & 40828 (502)                   & 0.351	& 0.141 & 0.974                \\ \hline
    \textit{Small}          & \textit{Female}                 & 9588 (133)                    & 0.154	& 0.054 & 0.989                \\
                            & \textit{Male}                   & 40828 (502)                   & 0.163	& 0.059 & 0.988                \\ \hline
    \textit{Medium**}       & \textit{Female}                 & 9588 (133)                    & 0.079	& 0.028 & 0.994                \\
                            & \textit{Male}                   & 40828 (502)                   & 0.088	& 0.032 & 0.994                \\ \hline
    \textit{Large**}        & \textit{Female}                 & 9588 (133)                    & 0.067	& 0.024 & 0.995                \\
                            & \textit{Male}                   & 40828 (502)                   & 0.074	& 0.027 & 0.995                \\
\end{tabular}
\caption{Weighted mean WER, CER, and BSS results for the gender attribute on the Common Voice data (** p < .01).}
\label{tab:whisper_gender_pre_trained}
\end{table}

Applying our fairness metric, the WER Parity, to our results, indicates that all models exhibit fair performance on the gender attribute indicating equal quality of service. In contrast, statistical testing shows that larger models exhibit significant gender 
biases. The medium and large models present relative gender bias of 0.9\% (11.4\%) and 0.7\% (10.5\%) on the weighted WER metric, respectively, in absolute terms with relative values between brackets. 
Thus, the larger the models get, the more predictive bias it exhibits. 

Despite these biases, the overall trend shows that increasing model size generally improves performance, as indicated by decreased weighted average WER and CER and increased BSS scores. However, the utility of the BSS metric is limited due to inconsistencies in its evaluation of semantic similarity and found to be not robust for single sentences and, thus, deemed irrelevant for the Common Voice data. This highlights the need for a consistent and accurate semantic metric. Moreover, the issue of "hallucinations" in smaller models, where the ASR system generates spurious transcriptions, was noted, especially for short or context-limited segments. The latter finding is corroborated by recent research, which indicates that hallucinations are more prevalent in segments with extended non-vocal durations \cite{koenecke2024careless}.

Inference times increase substantially with model size, suggesting a trade-off between performance and computational efficiency. The tiny model demonstrated the fastest processing time, with the base model requiring 1.1 times longer, the small model 1.7 times longer, the medium model 3.1 times longer, and the large model 4.4 times longer. This performance versus runtime trade-off suggests that the medium model may offer an optimal balance for practical applications.

\subsection{Dutch National Public Broadcasting organisation results}
The data set from the NPO, comprising real world radio and TV shows, provides a different challenge due to its longer and more varied audio segments. First, it is noted that the model speeds are in the same relative order as for the Common Voice data. Here, the medium model shows the lowest WER across speech categories, while the large model demonstrates fairness across all categories but performs similarly or worse than the tiny model, as shown in Table \ref{tab:npo_foundational_male_female_tv_shows}. This suggests that the large model might be too complex for this task, potentially overfitting to the training data.

\begin{table}[h!]
    \centering
    \begin{tabular}{l|l|l|l|l}
    \textbf{Model} & \textbf{ES (M, F)} & \textbf{NES (M, F)} & \textbf{Radio} (\textbf{M, F}) & \textbf{All (M, F)} \\ \hline
    Tiny    & .484** (.515, .424) & .684 (.668, .744) & .638 (.686, .466) $\neq$ & .597* (.624, .519) \\ \hline
    Base    & .356*** (.385, .3) $\neq$ & .593 (.575, .662) & .522 (.579, .323) $\neq$ & .485** (.515, .402) $\neq$ \\ \hline
    Small   & .261 (.285, .214) $\neq$ & .41 (.397, .459) & .409* (.442, .295) $\neq$ & .356 (.375, .3) $\neq$ \\ \hline
    Medium  & \textbf{.185} (\textbf{.199}, \textbf{.157}) $\neq$ & \textbf{.322} (\textbf{.312}, \textbf{.36}) & \textbf{.303} (\textbf{.331}, \textbf{.206}) $\neq$ & \textbf{.266} (\textbf{.282}, \textbf{.223}) $\neq$ \\ \hline
    Large  & .48* (.51, .421) & .837 (.82, .903) & .784*** (.803, .717) & .691 (.714, .626)     
    \end{tabular}
    \caption{Weighted mean WER results of the foundational Whisper models on the NPO data, split on male versus female groups and the different types of shows (* p < 0.05, ** p < 0.01, *** p < 0.001). Note that the unequal sign $\neq$ indicates unfairness as measured by means of our WER Parity metric. }
    \label{tab:npo_foundational_male_female_tv_shows}
\end{table}

The weighted mean CER results are shown in Table \ref{tab:npo_foundational_male_female_tv_shows_cer} and show the same trends as the WER scores. Again, values are substantially lower than WER scores, indicating lucky guesses, conjugations, or phonetically similar words. This finding adds to the argument for using CER as an additional metric to the WER and BSS.

\begin{table}[h!]
    \begin{tabular}{l|l|l|l|l}
    \textbf{Model} & \textbf{ES (M, F)}    & \textbf{NES (M, F)}   & \textbf{Radio (M, F)} & \textbf{All (M, F)}   \\ \hline
    Tiny           & .261, (.278, .229) & .439, (.42, .511)  & .411, (.443, .296) & .366, (.381, .321) \\ \hline
    Base           & .203, (.221, .168) & .421, (.4, .504)   & .343, (.383, .202) & .318, (.336, .266) \\ \hline
    Small          & .168, (.183, .138) & .301, (.29, .344)  & .297, (.32, .215)  & .251, (.265, .213) \\ \hline
    Medium         & \textbf{.129}, (\textbf{.14}, \textbf{.107})  & \textbf{.239}, (\textbf{.227}, \textbf{.286}) & \textbf{.226}, (\textbf{.249}, \textbf{.145}) & \textbf{.195}, (\textbf{.206}, \textbf{.164}) \\ \hline
    Large          & .414, (.44, .364)  & .734, (.716, .806) & .722, (.742, .651) & .615, (.635, .556)
    \end{tabular}
    \caption{Weighted mean CER results of the foundational Whisper models on the NPO data, split on male versus female groups and the different types of shows.}
    \label{tab:npo_foundational_male_female_tv_shows_cer}
\end{table}

Eloquent speech was better recognized than non-eloquent and radio speech, indicating the challenges of transcribing spontaneous speech. Female speech was generally better recognized than male speech, consistent with the Common Voice findings. Also similar to the Common Voice findings, the CER results follow the same trend as the WER metric. 

Statistical tests highlight gender biases in the NPO data. For example, the tiny and base models exhibited overall absolute gender biases of 10.5\% (20.2\%) and 11.3\% (28.1\%) weighted WER, respectively. Eloquent speech showed similar biases, with the large model presenting a gender bias of 8.9\% (21.1\%) weighted WER. Non-eloquent speech did not show gender bias, while the small and large models showed substantial biases on radio speech of 14.7\% (49.8\%) and 8.6\% (12\%), where that of the large model is similar to the Common Voice data. 

The WER Parity tests indicated more unfairness for the radio shows where only the large model is fair towards male and female instances. Moreover, eloquent shows note more unfairness, whereas non-eloquent shows seem fair for all model sizes. Purely based on fairness tests, the large model is the only fair model providing equal quality of service on all shows. Thus, neglecting overall performance, the large model is the fairest and seems the best option. Considering performance, the medium model outperforms the rest and could be considered the best.

\section{Discussion}
Generally, performance is substantially worse on the data of the Dutch National Public Broadcasting organisation in comparison to the Common Voice data. This can be related to the differences in domains of read speech versus spontaneous speech. Spontaneous speech is more difficult for ASR systems since it can contain filled pauses, restarts of words or phrases, interjections, ungrammatical constructions, or even unknown or mispronounced words \cite{ward1989understanding}. Manual inspection of the NPO transcriptions yielded multiple issues, highlighting the need for future curation of this data and a reason for possible worse results. These issues include empty or timeless segments, inconsistent transcription of numbers and music, and removal of arbitrary non-word sounds from transcriptions.

Our results are in line with previous research, showing better recognition of female speech over male speech. 
The trends in WER and CER scores are consistent, leading us to primarily rely on WER due to its widespread adoption and relevance in evaluating ASR systems. The BSS metric turned out to be inconsistent and unreliable and was not further taken into account. Larger models, while offering better overall accuracy, display more predictive biases, underscoring a trade-off between performance efficiency and computational cost. This trade-off is particularly crucial in real-world applications where computational resources and response time are limiting factors.

The analysis of the NPO dataset, comprising diverse TV and radio broadcasts, reveals that eloquent speech typically achieves the best recognition performance, followed closely by radio and non-eloquent speech. Interestingly, the medium-sized Whisper model balances performance and computational efficiency most effectively across different speech types. This finding underscores the complexities of developing ASR models that perform equitably across varied speech contexts.

The comprehensive benchmarking of Whisper models across the Common Voice and the NPO data sets reveals several key insights:

\begin{itemize}
    \item \textit{Model Performance}: Larger models generally offer better performance in terms of WER, CER, and BSS but come with increased computational costs.
    \item \textit{Bias and Fairness}: Significant biases were observed across gender 
    attributes, with larger models exhibiting more pronounced disparities. This highlights the need for addressing these biases, for instance, through targeted fine-tuning.
    \item \textit{Contextual Challenges}: The type of speech (eloquent vs. non-eloquent) and the format of the data (short-form vs. long-form) significantly impact ASR performance, indicating the importance of context in model training and evaluation.
    \item \textit{Metric Limitations}: The BSS metric, while useful, showed limitations in evaluating semantic similarity, suggesting a need for alternative metrics or adjustments in its application.
    \item \textit{Practical Considerations}: The trade-off between model performance and inference time is crucial, with the medium model emerging as a balanced choice for many applications.
\end{itemize}

This analysis underscores the complexities in developing fair and accurate ASR systems and highlights the necessity for continued research to mitigate biases and improve the performance of these systems in diverse real-world scenarios. 

The practical implications of these findings are substantial. They reveal a clear trade-off between model performance, demographic bias, and operational costs, which stakeholders must carefully consider in different application contexts. For instance, the use of Whisper ASR models for Dutch language processing in media or educational settings must take into account the potential for quality-of-service harms, especially if certain demographic groups are systematically disadvantaged by bias in subtitle accuracy. These biases are not merely technical challenges; they pose real ethical concerns, especially given the growing dependence on automatic subtitling in media and educational content.

\subsection{Limitations and Future Directions}
Despite the strengths of our approach, several limitations must be acknowledged:
\begin{itemize}
    \item \textit{Data Quality and Representation}. Our study assumes the demographic information in the Common Voice dataset is accurate, but inaccuracies in these self-reported data could skew results. Additionally, the NPO data had issues like empty segments and inconsistent transcriptions, possibly impacting validity of results. Future work should improve data quality verification and standardize text normalization.    
    \item \textit{Generalizability of Results}. While we have provided extensive analysis within the context of Dutch language data, the generalizability of our findings to other languages or dialects may be limited. Additionally, Whisper's non-deterministic nature introduces variability, affecting result generalizability. Further studies are needed to explore how these models perform across a broader spectrum of languages, accents, age groups, and domains.
    \item \textit{Semantic Similarity Challenges}. The BERT-based semantic similarity metric did not perform as expected in our framework, indicating the need for further refinement and testing of semantic analysis tools in ASR evaluations.
    \item \textit{Model Interpretation}. The black-box nature of Whisper models limits robust conclusions about biases. Future work should focus on model interpretation for deeper insights.
    \item \textit{Additional Considerations}. Including new Whisper models in future comparisons could provide more insights.
\end{itemize}

Future research should focus on enhancing data quality, improving the robustness of semantic similarity measures, and exploring the impact of fine-tuning ASR models on reducing demographic biases. Additionally, analyzing Whisper's variability and expanding the scope of studies to include more diverse languages, dialects, and domains (e.g., healthcare, customer service, legal) will help improve the generalizability of findings. Finally, model interpretation could yield valuable conclusions about the model's systematic behavior, helping to validate conclusions on biases, while new and larger Whisper models could be included for more insights.

The predictive biases observed in this study highlight the necessity for ongoing research into ASR fairness. Future work should explore the effectiveness of fine-tuning strategies to mitigate biases and enhance overall performance while ensuring fairness across all user groups. Additionally, researchers should continually reassess the moral frameworks and fairness metrics used to ensure they remain relevant and effective in new contexts.

\section{Conclusion}
This study evaluated five state-of-the-art Whisper ASR systems for demographic predictive bias on the Dutch language, focusing on quantifying gender bias through a morally just framework that centers on quality-of-service harms in automatic subtitling. Using the moral justification framework by Weerts et al. \cite{weerts2022does}, we proposed a WER Parity metric, which enhances our capability to reason about quality-of-service harms among gender groups. This approach aligns with the guidelines proposed by Blodgett et al., who have urged researchers to provide explicit statements justifying why certain system behaviors that manifest as "bias" are considered harmful and to whom these biases are detrimental \cite{blodgett2020language}.

Our study integrated WER and CER metrics, supplemented by a BERT-based semantic similarity measure, to provide a comprehensive assessment of ASR performance. We carefully checked for normality and homogeneity of variances, using robust non-parametric tests when necessary. This ensures that our conclusions about demographic biases in ASR performance are well-grounded.

Five different Whisper models have been tested for their performance in terms of word and character error rate, a semantic similarity metric, and a WER-based fairness metric. Larger models demonstrated better performance but also more pronounced biases and higher operational costs. Gender-based performance differences were notable, with female speech receiving better recognition than male speech across various settings. These insights highlight the ongoing challenges in creating fair ASR systems, providing valuable guidance for developing more inclusive and ethically sound technologies.

This study not only provides insights into the demographic biases present in foundational Whisper models but also contributes to a broader understanding of how technological advancements can be aligned with principles of fairness and equity. By employing a more rigorous statistical framework and a morally grounded evaluation of ASR systems predictive biases, our work contributes to a more reliable, fair, and equitable ASR technologies, reinforcing the belief that "Everyone Deserves Their Voice to Be Heard."

\begin{credits}
\subsubsection{\ackname} This work has been partially supported by the KOIOS project (Grant Agreement Nº 101103770), executed at the Netherlands Organisation for Applied Scientific Research (TNO) with additional data resources provided by the Dutch Public Broadcasting organization, the Nederlandse Publieke Omroep (Dutch Foundation for Public Broadcasting or NPO\footnote{https://npo.nl/}). We would especially like to thank Egon Verharen for his continuous support and valuable feedback during the project. 


\subsubsection{\discintname}
The authors have no competing interests to declare that are
relevant to the content of this article.

\end{credits}
%
%
%
\bibliographystyle{splncs04}
\bibliography{paper}
\end{document}